%% file: main.tex
\documentclass[10pt,twocolumn,letterpaper]{article}

\usepackage{iccv}              
\usepackage{hhline}
\usepackage{algorithm}%
\usepackage{algorithmicx}%
\usepackage{algpseudocode}%
\usepackage{listings}%
\usepackage{dsfont}
\usepackage{hhline}


\definecolor{iccvblue}{rgb}{0.21,0.49,0.74}
\usepackage[pagebackref,breaklinks,colorlinks,allcolors=iccvblue]{hyperref}

\def\paperID{14} 
\def\confName{ICCV}
\def\confYear{2025}

\title{AttZoom: Attention Zoom for Better Visual Features}

\author{Daniel DeAlcala, Aythami Morales, Julian Fierrez, Ruben Tolosana \\
Biometrics and Data Pattern Analytics Lab, Universidad Autonoma de Madrid, Spain\\
{ \tt\small daniel.dealcala@uam.es} {\tt\small aythami.morales@uam.es} {\tt\small julian.fierrez@uam.es} {\tt\small ruben.tolosana@uam.es \tt\small  }}

\begin{document}
\maketitle
\input{sec/0_abstract}    
\input{sec/1_intro}
\input{sec/2_related}
\input{sec/3_method}
\input{sec/4_integration}
\input{sec/5_expprot}
\input{sec/6_results}

\input{sec/7_conclusions}
\input{sec/8_acks}

{
    \small
    \bibliographystyle{ieeenat_fullname}
    \bibliography{main}
}

\end{document}

%% file: sec/0_abstract.tex
\begin{abstract}
We present Attention Zoom, a modular and model-agnostic spatial attention mechanism designed to improve feature extraction in convolutional neural networks (CNNs). Unlike traditional attention approaches that require architecture-specific integration, our method introduces a standalone layer that spatially emphasizes high-importance regions in the input. We evaluated Attention Zoom on multiple CNN backbones using CIFAR-100 and TinyImageNet, showing consistent improvements in Top-1 and Top-5 classification accuracy. Visual analyses using Grad-CAM and spatial warping reveal that our method encourages fine-grained and diverse attention patterns. Our results confirm the effectiveness and generality of the proposed layer for improving CCNs with minimal architectural overhead.
\end{abstract}

%% file: sec/1_intro.tex
\section{Introduction}
\label{sec1}

The attention mechanisms in AI allow models to focus their capacity on the most relevant information, much as the human brain allocates cognitive resources to different aspects of sensory input through attention \cite{anderson2005directed}. This ability is crucial for efficient information processing in many human activities such as learning \cite{becerra24eccvw,daza24att}, as not all data are equally important. A key trend in AI has been to improve performance by drawing inspiration from brain function, leading to the incorporation of attention mechanisms into AI models in what is called brain-inspired artificial intelligence (BIAI) \cite{yang2018survey, ren2024brain}. These mechanisms help systems filter out irrelevant details and focus on critical aspects, improving tasks such as pattern recognition and decision-making \cite{lai2020understanding}.

Deep convolutional neural networks (CNNs) have become the backbone of many state-of-the-art image recognition systems \cite{camacho22cnns}. However, standard CNNs process all spatial regions in an input image uniformly, which can lead to suboptimal performance when only certain regions are relevant for a given task \cite{woo2018cbam}. This is common in many image-based recognition problems, e.g., in biometrics for person recognition \cite{tome13regions,tome15fusing,ester18regions,tolo22fakes}. To address this non-uniformity in the relevance of different parts of the data, some researchers have developed weight-based fusion schemes \cite{fierrez06phd,fierrez18fusion1}, combining the different pieces of information with fixed or adaptive weights \cite{fierrez05score,fierrez05adapted}. In a different line of action, attention mechanisms have been proposed to guide the focus of trained networks toward the most informative parts of the image~\cite{mnih2014recurrent}.

In the literature, the development of attention mechanisms has played a crucial role in computer vision. Most attention methods rely on dedicated attention blocks integrated throughout the network architecture, leading to models such as \cite{woo2018cbam,hu2018squeeze,xie2017aggregated,liu2022convnet, HASSANIN2024102417}. In contrast, our approach introduces a standalone attention layer that helps the model focus on different regions of the image. Its independent nature makes it easy to implement and apply, while allowing seamless integration into any model architecture. This wrapping of the existing model is inspired by our previous work \cite{morales21snets}, where new layers were introduced not to emphasize certain regions, but to remove certain sensitive data.

We evaluated our approach by inserting the proposed attention layer into several popular CNN backbones and performing classification experiments on CIFAR-100 \cite{krizhevsky2009learning} and TinyImageNet \cite{chrabaszcz2017downsampled} datasets. The results show consistent improvements across all models tested.

Our contributions are threefold:

\begin{itemize}
    \item We propose an attention mechanism that enhances feature extraction from images, easily applicable without architectural changes to subsequent processing via existing neural models and new ones.
    \item We integrate the attention module into multiple CNN backbones without architecture-specific modifications.
    \item We demonstrate the effectiveness of our method through extensive experiments on benchmark datasets.
\end{itemize}

The remainder of this paper is organized as follows. Section~\ref{sec2} reviews related work on attention mechanisms. Section~\ref{sec3} introduces our proposed attention method, AttZoom, including its mathematical formulation. Section~\ref{methodintegration} describes the integration process into existing architectures, along with pseudocode. Section~\ref{exppro} outlines our experimental setup, covering datasets, models, and training details. The results are presented in Section~\ref{Sec:Results}, followed by the conclusions in Section~\ref{sec:conclusions}.

%% file: sec/2_related.tex
\section{Related Works}
\label{sec2}

Attention mechanisms have been extensively studied in the literature, especially in recent years, leading to significant advances in AI. These methods can be categorized in various ways. In this work, we focus on the mechanisms of attention applied to images and classify them into three main groups: channel attention, spatial attention, and self-attention, following the taxonomy presented in \cite{HASSANIN2024102417, guo2022attention}.

\subsection{Channel Attention}

Channel attention mechanisms learn to recalibrate the importance of each channel within a convolutional feature map. In convolutional networks, each channel represents a specific set of features detected by different filters. Channel attention enables the network to selectively emphasize or suppress certain channels based on the image context, improving the representation of relevant features \cite{fierrez18fusion2}.

One of the most fundamental works in channel attention is Squeeze \& Excitation (SE) \cite{hu2018squeeze}. The SE block compresses each channel into a single value using global average pooling (Squeeze) and then adjusts its importance through two fully connected layers (Excitation): one with ReLU and another with Sigmoid. Several works have aimed to improve SE efficiency, such as Efficient Channel Attention (ECA) \cite{Wang_2020_CVPR}, which replaces fully connected layers with 1D convolutions to eliminate the need for dimensionality reduction. Similarly, \cite{yang2020gated} applies L2 normalization per channel, scales with a learnable vector, uses tanh normalization \cite{fierrez05score} for attention, and integrates the output with the input via a residual connection. Another notable work is ResNeSt \cite{Zhang_2022_CVPR}, a ResNet variant that incorporates split-attention blocks, leveraging global pooling, convolutions, and softmax to adjust channel importance. Furthermore, CBAM \cite{woo2018cbam} and BAM \cite{park2018bam} combine channel and spatial attention by extracting descriptors through maximum and average pooling, processing them with a multilayer perceptron, and applying a sigmoid activation to generate an attention map.

A wide range of other studies have explored channel attention. For example, \cite{dai2019second} employs covariance matrices decomposed into eigenvalues to construct its attention mechanism, while \cite{qin2021fcanet} uses global average pooling in the frequency domain. For a more comprehensive overview of channel attention methods, we refer readers to extensive surveys such as \cite{HASSANIN2024102417, survey3, guo2022attention}.

\subsection{Spatial Attention}

Spatial attention methods highlight specific regions within an image by assigning weights to different spatial locations in the feature map. Unlike channel attention, which modulates entire feature channels, spatial attention operates directly on the spatial distribution of features. Our approach falls into this category as it emphasizes the importance of particular image regions to enhance recognition performance.

CBAM \cite{woo2018cbam} combines channel and spatial attention by taking advantage of both average-pooled and max-pooled features to generate a 2D spatial attention map. The Recurrent Attention Model (RAM) \cite{mnih2014recurrent} employs recurrent neural networks (RNNs) and reinforcement learning (RL) to train the model in selecting relevant regions for attention. Several other works also build spatial attention mechanisms based on RNNs \cite{gregor2015draw,xu2015show,ba2014multiple}. Co-attention \& Co-excitation \cite{hsieh2019one} utilizes non-local networks to capture long-range dependencies.

A significant portion of spatial attention research is based on pyramids \cite{hu2020span,li2020spatial,zhao2019pyramid}, where attention maps are generated from features extracted at different levels of the model. For a more detailed discussion of spatial attention methods, we refer the reader to comprehensive surveys such as \cite{HASSANIN2024102417, survey3, guo2022attention}.

\subsection{Self Attention}

Self attention enables each position in a data sequence (image, text, etc.) to interact with all other positions, assigning weights based on their relevance. This allows for more effective long-range dependency modeling compared to traditional convolutions, which have a more limited receptive field. Self-attention is the core mechanism behind Transformers, which have not only revolutionized NLP \cite{vaswani2017attention, devlin2019bert, daietal2019transformer, choromanski2020rethinking,mancera2025pba} but also achieved significant advances in computer vision \cite{zhu2020deformable, chen2020generative, carion2020end,sergio25cvprw} and pattern recognition on temporal functions \cite{paula23gait,paula24swipe}.

Self attention can be traced back to the LSTMs \cite{hochreiter1997long}, where an attention vector was generated at each hidden state to indicate which parts of the sequence should interact \cite{tolo21aaai}. However, Self-Attention gained prominence due to Transformers \cite{vaswani2017attention}. Transformers follow an encoder-decoder architecture, where the encoder applies a self-attention mechanism to capture global dependencies in the input, followed by a feedforward layer. The decoder has a similar structure but incorporates a cross-attention layer, allowing it to focus on the relevant information from the encoder. Numerous Transformer variants have been developed for specialized tasks. For example, Swin Transformers \cite{liu2021swin} introduce sliding window attention and hierarchical structures to enhance performance in computer vision tasks.

%% file: sec/3_method.tex
\section{Proposal: Attention Zoom (AttZoom)}\label{sec3}

\begin{figure*}[t!]
\centering
\includegraphics[width=0.99\linewidth]{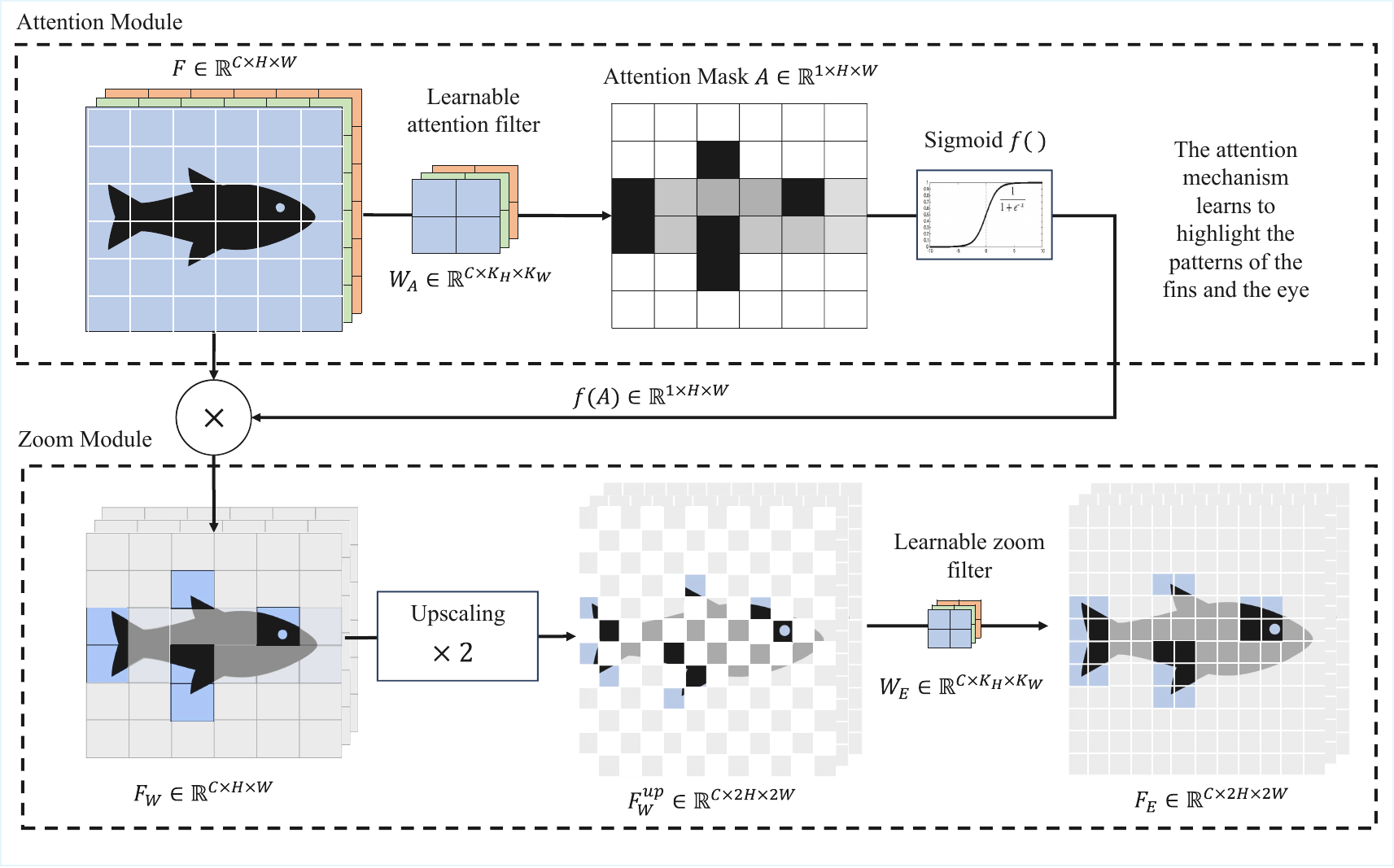} 
\caption{Block diagram of the proposed attention mechanism. Note that this is a conceptual representation, and the intermediate feature $F$ does not necessarily correspond to an image, but rather to a learned representation. In this example, the most relevant features are located around the fins and the eye. The attention module learns to mask the most relevant regions, while the zoom module enhances the patterns within these regions.}
\label{block_diagram}
\end{figure*}

In this work, we introduce our method, Attention Zoom, a spatial attention mechanism that incorporates Attention-Zoom layers. Intuitively, these layers help the model focus on the most important regions of the image by amplifying relevant information for classification while suppressing less informative areas that the network deems irrelevant. In Fig. \ref{block_diagram}, we show the block diagram of the proposed attention/enhancement mechanism.

We start with an intermediate feature map $F \in \mathbb{R}^{C\times H \times W}$. The first step involves learning a Spatial Attention Map $A\in \mathbb{R}^{1\times H \times W}$. To obtain this Spatial Attention Map, the model learns a convolutional kernel that processes the intermediate feature map $F$ to identify the most relevant regions. We denote this convolutional kernel by $W_A \in \mathbb{R}^{C\times K_H \times K_W}$. By convolving this kernel with the feature map, we obtain our attention map: 
\[A = W_A * F\]

To preserve spatial dimensions $H\times W$, we use a stride of 1 and zero padding. Once the Attention Map is obtained, we apply the following function:
\[
f(A) = 
\begin{cases} 
1, & \text{if } \sigma(A) \geq \textrm{threshold} \\
\sigma(A), & \text{if } \sigma(A) < \textrm{threshold}
\end{cases}
\]

\[
\sigma(A) = \frac{1}{1 + e^{-A}}
\]

\noindent This step is crucial because it allows the network to preserve the regions it considers important while gradually suppressing irrelevant areas. We obtain the weighted feature map by performing an element-wise multiplication $\odot$ between the input feature map (each channel independently) and the learned attention map:

\[
F_W = F \odot f(A), \quad \text{with } f(A) \in \mathbb{R}^{1 \times H \times W}
\]


After the weighted feature map $F_W$ we proceed to the `zoom' operation. This step amplifies the information in the regions deemed important by the attention map while diluting or reducing the details in less relevant areas.

To achieve this, we apply two transformations. The first involves inserting a zero into every spatial position of $F_W$ effectively increasing the resolution. This produces the Upsampled Weighted Feature Map:

\[F_W^{Up} = \textrm{Upsample}(F_W) \in  \mathbb{R}^{C \times H' \times W'}\]

\noindent where the new dimensions are defined as:

\[H' = 2H, W' =2W\]

\noindent (Note that, without loss of generality, the multiplier introduced here $2$ can be increased for a higher zoom.)

This upsampling operation is illustrated in Figure \ref{block_diagram}. Mathematically, it can be expressed as follows:

\[
F_W^{Up}(c, 2h, 2w) = F_W(c, h, w), \quad
F_W^{Up}(c, 2h+1, w) = 0, \quad \]
\[
F_W^{Up}(c, h, 2w+1) = 0, \quad
F_W^{Up}(c, 2h+1, 2w+1) = 0
\]

\noindent for $c=0,\ldots,C-1; h=0,\ldots,H-1; w=0,\ldots,W-1$.

We then apply a convolution, obtaining the Enhanced Feature Map:

\[F_E=W_{E} * F_W^{Up}\] 

This convolution modulates the information density, either amplifying or suppressing features depending on their importance as determined by the attention map.
\begin{itemize}
    \item In regions where $f(A)$ reduced values (i.e., where the attention map multiplied the pixels to a value below a threshold), inserting zeros and applying a convolution further diminishes this information, pushing values closer to zero.
    \item In contrast, in regions where $f(A)$ kept values unchanged (ie, where $A$ had values above a threshold), inserting zeros followed by convolution performs a local up-sampling. This effectively expands the feature representation in those areas, increasing the amount of retained information and enlarging its receptive field.
\end{itemize}

This process ensures that important features are emphasized while irrelevant regions fade out, allowing the model to focus on the most discriminative parts of the image.

%% file: sec/4_integration.tex
\section{Method Integration} \label{methodintegration}

Before proceeding with our experiments, we highlight several key aspects of the Attention-Zoom layers and how they differ from existing attention-based methods.


Most attention mechanisms introduce an attention block at multiple points within a model, effectively defining a new architecture (e.g., ResNext \cite{xie2017aggregated}, SE \cite{hu2018squeeze}, CBAM \cite{woo2018cbam}). These architectures are typically modifications of a backbone such as ResNet \cite{he2016deep}, incorporating attention-based enhancements to improve performance. Other approaches leverage attention blocks to design entirely new architectures, as seen in Transformer-based models \cite{vaswani2017attention}.

In contrast, our work introduces the Attention-Zoom layer, which expands the information and increases the receptive field in certain image regions while reducing the significance of others. Unlike traditional attention modules, Attention-Zoom is not structured as a block and does not modify the network architecture. Instead, it is a modular layer that can be seamlessly integrated at any stage of a model.

This design offers several advantages. First, it maintains the original network architecture while adding only a few extra parameters. Second, it is highly adaptable and can be incorporated into any model without requiring structural modifications (simply by adding the layer in the early stages). Moreover, it can be combined with other attention-based approaches such as ResNext \cite{xie2017aggregated}, SE \cite{hu2018squeeze}, CBAM \cite{woo2018cbam}, etc. The pseudocode for the Attention-Zoom layer is presented in Algorithm \ref{alg:attention_zoom}. 

\begin{algorithm}[t]
\caption{Attention-Zoom Layer}\label{alg:attention_zoom}
\begin{algorithmic}[1]
    \State \textbf{Input:} Feature Map $F \in \mathbb{R}^{C \times H \times W}$ 
    \State \textbf{Output:} Enhanced Feature Map $F_E \in \mathbb{R}^{C \times H' \times W'}$ 
    
    \State \textbf{Step 1: Compute Spatial Attention Map} 
    \State $A = \sigma(W_A * F)$  \hfill \textit{(Conv + Sigmoid)}

    \State \textbf{Step 2: Apply Attention Mask}
    \State $A = \mathbf{1}(A \geq threshold) + A \cdot \mathbf{1}(A < threshold)$
    \State $F_W = F \odot A$  \hfill \textit{(Element-wise multiplication)}

    \State \textbf{Step 3: Apply Upsampling with Zeros}
    \State $F_W^{Up} = \text{UpsampleZeros}(F_W)$

    \State \textbf{Step 4: Apply Enhancement Convolution}
    \State $F_E = W_E * F_W^{Up}$  \hfill \textit{(Final convolution)}

    \State \textbf{Return} $F_E$
\end{algorithmic}
\end{algorithm}

%% file: sec/5_expprot.tex
\section{Experimental Protocol} \label{exppro}

In this work, we integrate the Attention-Zoom layers into various architectures trained on different datasets under a well-defined experimental protocol that ensures fair comparisons.

\subsection{Data and Models} \label{dataandmodels}

We evaluated our method on two widely used benchmark datasets. Specifically, we consider CIFAR-100 \cite{krizhevsky2009learning} and Tiny-ImageNet \cite{chrabaszcz2017downsampled}, as they provide a suitable balance between complexity and scalability for image classification tasks. 

We conducted experiments on two categories of architectures. First, we evaluate Attention-Zoom layers on standard state-of-the-art models widely used in the literature, including ResNet-50 \cite{he2016deep}, DenseNet-121 \cite{huang2017densely} and MobileNet \cite{howard2017mobilenets}. 

Second, we evaluate the integration of our method into architectures that already incorporate attention mechanisms, specifically Squeeze-and-Excitation (SE) \cite{hu2018squeeze}, Convolutional Block Attention Module (CBAM) \cite{woo2018cbam}, and ResNeXt \cite{xie2017aggregated}. As these attention mechanisms have been integrated into ResNet-50, they provide particularly meaningful comparisons for our study.

For each architecture, we follow a two-step evaluation: (1) training the model without Attention-Zoom layers, and (2) retraining it with Attention-Zoom layers, which allows us to directly compare their impact on performance.

\subsection{Collaboration, Not Competition} \label{ColNoCom}

The primary goal of our study is to evaluate whether the inclusion of Attention-Zoom layers effectively enhances model performance. As discussed in Sect. \ref{methodintegration}, a key advantage of our method is its modularity: it consists of a single layer that can be seamlessly integrated into any model.

Therefore, while direct comparisons with attention-based methods are relevant, our approach extends beyond competition. Instead, we demonstrate how Attention-Zoom layers can be incorporated into existing attention-based architectures to further improve their performance. This highlights the flexibility and complementary nature of our method.

\subsection{Model Training}

To ensure a rigorous and fair evaluation, we adopt a robust hyperparameter optimization strategy. Specifically, we utilize Optuna \cite{optuna_2019}, a Bayesian optimization framework, to perform a hyperparameter search.

For each result reported in Sec. \ref{Sec:Results}, we conduct an Optuna Bayesian search over 30 trials. This means that each model undergoes 30 training runs with different hyperparameter configurations, following a guided search strategy. The best configuration is then selected for comparison.

This approach ensures that our reported results are not biased by stochastic factors such as random hyperparameter selection. Instead, they reflect the best achievable performance for each model, making our comparisons statistically meaningful.

The following parameters are optimized by Bayesian search:

\begin{itemize} 
\item \textbf{Batch Size:} 32, 64, 128, or 256. 
\item \textbf{Learning Rate:} Sampled logarithmically between $10^{-5}$ and $10^{-1}$. 
\item \textbf{Weight Decay:} Sampled logarithmically between $10^{-6}$ and $10^{-2}$. 
\item \textbf{Optimizer:} Adam or SGD. For SGD, we tuned the momentum between 0.7 and 1 in linear scale. 
\item \textbf{Scheduler:} One of CosineAnnealingLR, StepLR, ReduceLROnPlateau, or OneCycleLR. For StepLR, we also tune the step size between 10 and 30 (linear scale). 
\end{itemize}

Models are trained from scratch for 50 epochs with early stopping (patience of 5 epochs). The data augmentation strategy includes RandomCrop, RandomHorizontalFlip, and AutoAugment. 

While more complex augmentation, optimization strategies, or pre-trained models could be employed, our setup is sufficiently robust to achieve competitive results under realistic training conditions. We intentionally avoid overly specialized training techniques or pre-trained models, as the goal of this work is not to extract marginal gains for each model but to establish a strong and fair experimental framework for comparing models trained with the Attention-Zoom Layer.

%% file: sec/6_results.tex
\section{Results} \label{Sec:Results}

We first report in Table \ref{tab:AZCIFAR} the Top-1 and Top-5 accuracy scores for the models presented in Sec. \ref{dataandmodels}, all trained on CIFAR-100. In addition to the results, we include the best hyperparameters identified by Optuna \cite{optuna_2019} using Bayesian search to achieve these performances. We observe that our method consistently improves the performance of all evaluated models. Although some architectures such as DenseNet121 and SE ResNet exhibit modest gains of less than 1\%, others such as MobileNet and ResNet50 achieve improvements of around 10\%. The remaining models fall somewhere in between.

\begin{table*}[]
\resizebox{\textwidth}{!}{%
\begin{tabular}{|l|c|c|c|}
\hline
\bf Architecture             & \bf Top-1 Acc (\%) & \bf Top-5 Acc (\%) & \bf Hyper-params \\  \hhline{|=|=|=|=|}
ResNet50                 &       58.50         &       87.75         &     128, $1^{-3} $, $3^{-4}$, Adam, OneCycleLR         \\
ResNet50 +AttZoom        &        67.66        &       90.15         &      128, $7^{-4}$, $1^{-4}$, Adam, OneCycleLR \\ \hline
MobileNet                &      63.62          &       88.61         &     64, $3^{-3}$, $2^{-5}$, Adam, OneCycleLR          \\
MobileNet +AttZoom       &      76.43          &       95.01         &     256, $7^{-3}$, $2^{-5}$, Adam, OneCycleLR         \\ \hline
DenseNet121              &        75.43        &       93.69         &      64, $3^{-4}$, $1^{-6}$, Adam, CosineAnnealingLR       \\
DenseNet121 + AttZoom    &    75.82            &       94.41         &      128, $2^{-4}$, $4^{-6}$, Adam,  CosineAnnealingLR     \\ \hline \hline
\bf Attention Architecture             & \bf Top-1 Acc (\%) & \bf Top-5 Acc (\%) & \bf Hyper-params \\ \hhline{|=|=|=|=|}
SE ResNet            &     76.95           &       94.19         &     64, $4^{-4}$, $2^{-5}$, Adam,  CosineAnnealingLR        \\ 
SE ResNet + AttZoom  &     77.06           &       94.38         &       64, $2^{-3}$, $6^{-4}$, SGD(0.99), CosineAnnealingLR       \\ \hline
CBAM ResNet           &     59.72           &       85.58         &       64, $5^{-4}$, $2^{-4}$, Adam,  OneCycleLR      \\
CBAM ResNet + AttZoom &      64.32         &      87.70          &      128, $5^{-4}$, $4^{-5}$, Adam, StepLR(29)        \\ \hline
ResNext                  &       72.83         &       92.59         &       64, $7^{-4}$, $1^{-5}$, Adam, ReduceLROnPlateau       \\
ResNext + AttZoom        &       77.53         &       95.34         &       64, $1^{-4}$, $4^{-3}$, Adam, CosineAnnealingLR       \\ \hline
\end{tabular}%
}
\caption{Top-1 and Top-5 accuracy on CIFAR-100 for all models introduced in Sec.~\ref{dataandmodels}, with and without AttZoom layers. The table also reports the best hyperparameters found via Bayesian optimization using Optuna~\cite{optuna_2019}.}
\label{tab:AZCIFAR}
\end{table*}

Table \ref{tab:AZIMAGENET} reports again the best results obtained through Bayesian optimization, this time on the Tiny-ImageNet dataset. The trends are consistent with those observed in the previous table: some models benefit substantially more from the introduction of the Attention-Zoom layers than others. In particular, SE ResNet and ResNeXt show comparable performance with and without our layer. However, models such as ResNet50 and MobileNet exhibit more significant performance improvements.

\begin{table*}[]
\resizebox{\textwidth}{!}{%
\begin{tabular}{|l|c|c|c|}
\hline
\bf Architecture             & \bf Top-1 Acc (\%) & \bf Top-5 Acc (\%) & \bf Hyper-params \\ \hhline{|=|=|=|=|}
ResNet50                 &       48.58         &       74.33         &    64, $2^{-3}$, $8^{-4}$, SGD(0.96),  ReduceLROnPlateau      \\ 
ResNet50 +AttZoom        &       63.23         &       84.12         &   32, $5^{-2}$, $2^{-4}$, SGD(0.9),  CosineAnnealingLR  \\ \hline
MobileNet                &       55.26         &       78.26         &  64, $1^{-2}$, $1^{-6}$, Adam, OneCycleLR            \\
MobileNet +AttZoom       &       64.89         &       85.30         &  64, $1{^-1}$, $1{^-4}$, SGD(0.8), CosineAnnealingLR            \\ \hline
DenseNet121              &       64.19         &       85.44         &  32, $6^{-5}$, $2^{-4}$, Adam, StepLR(30)           \\
DenseNet121 + AttZoom    &       66.29         &       86.45         &  64, $1{^-3}$, ${1^-3}$, SGD(0.9), CosineAnnealingLR       \\ \hline \hline
\bf Attention Architecture             & \bf Top-1 Acc (\%) & \bf Top-5 Acc (\%) & \bf Hyper-params \\ \hhline{|=|=|=|=|}
SE ResNet            &       65.92         &       86.01         &     64, $6^{-4}$, $2^{-4}$, Adam,  CosineAnnealingLR       \\ 
SE ResNet + AttZoom  &       66.06         &       86.36         &     64, $1^{-3}$, $5^{-2}$, Adam,  CosineAnnealingLR        \\ \hline
CBAM ResNet          &       52.38         &      76.82          &    64, $5^{-3}$, $6^{-4}$, SGD(0.9), StepLR(30)         \\
CBAM ResNet + AttZoom &      57.88         &      81.08          &    128, $2^{-2}$, $3^{-3}$, SGD(0.7), CosineAnnealingLR         \\ \hline
ResNext                  &       63.20         &      84.08          &    64, $2^{-5}$, $9^{-5}$, Adam, StepLR(30)        \\
ResNext + AttZoom        &       63.77         &      84.23          &    32, $2^{-3}$, $2^{-5}$, SGD(0.9), CosineAnnealingLR          \\ \hline
\end{tabular}%
}
\caption{Top-1 and Top-5 accuracy on Tiny-ImageNet for all models, with and without AttZoom layers. We also report the best hyperparameters found via Bayesian optimization using Optuna~\cite{optuna_2019}.}
\label{tab:AZIMAGENET}
\end{table*}

\begin{figure*}[t!]
\centering
\includegraphics[width=0.99\linewidth]{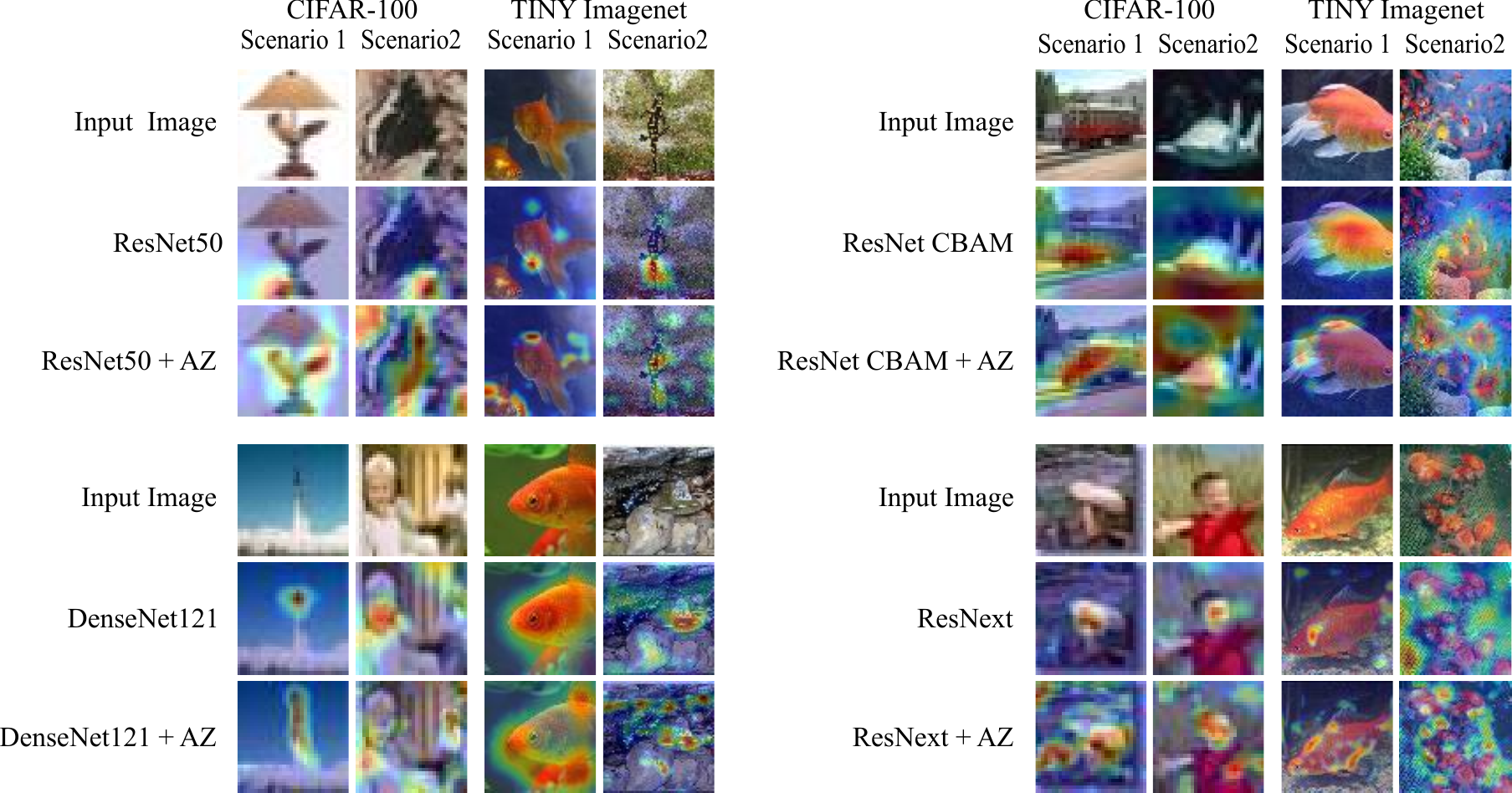} 
\caption{Grad-CAM visualizations for models with and without AttZoom layers. We show examples from four representative architectures (ResNet50, CBAM ResNet, DenseNet121, ResNeXt) on different datasets. Each row corresponds to either Scenario 1 (both models classify correctly) or Scenario 2 (both models fail) and both datasets.}
\label{GradCam}
\end{figure*}

\begin{figure*}[t!]
\centering
\includegraphics[width=0.99\linewidth]{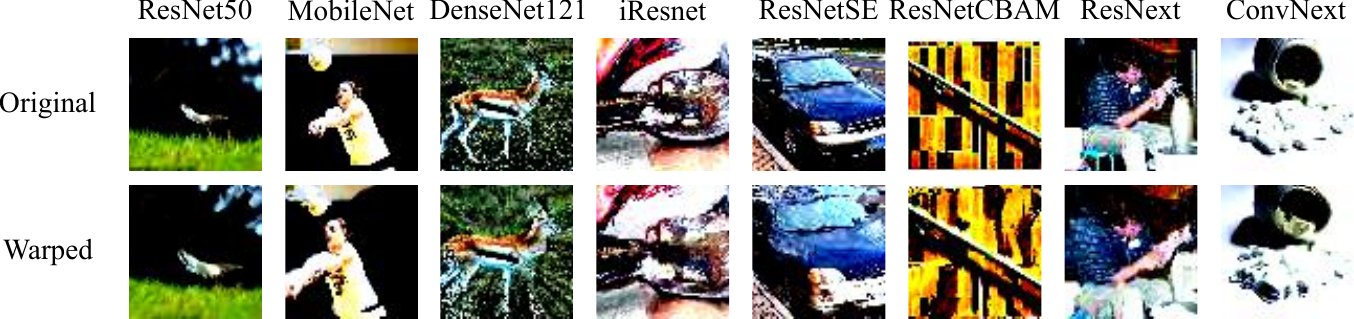} 
\caption{Warped image visualizations based on the attention produced by the AttZoom Layer. We show one correctly classified image per model, using the Tiny ImageNet dataset. The warping expands regions with high attention and contracts those with low attention. }
\label{warped}
\end{figure*}

An interesting observation arises when comparing models that already incorporate attention mechanisms with those that do not. For models with existing spatial attention modules (last 6 rows in Tables \ref{tab:AZCIFAR} \ref{tab:AZIMAGENET}), the performance gains tend to be smaller. However, our method still yields improvements, reaching 5\% in some cases. These results, along with the ease of implementation of our method (which consists of a single, easily integrable layer), suggest that this module can benefit a wide range of models, regardless of whether they already include attention mechanisms, but with even greater benefits observed in those that do not.

Given the simplicity of implementing our method, the goal is not to compete with existing spatial attention mechanisms, but rather to complement them (Sec. \ref{ColNoCom}). However, since the three evaluated attention-based architectures (S\&E ResNet, CBAM ResNet, and ResNeXt) are all built on the ResNet backbone, we compare these architectures (without our AttZoom) to a standard ResNet with AttZoom. This allows us to evaluate whether simply adding our layer to a standard ResNet can approach the performance of more complex architectures like the ones mentioned above. 

Our results show that simply adding our AttZoom layer brings the performance of standard ResNet close to—or even above—that of more complex attention-based models, as observed in cases like CBAM on TinyImageNet and CIFAR-100, and SE on TinyImagenet.

\subsection{Interpretability: GradCam Visualization}

To qualitatively analyze our Attention Zoom Layers, we employ the well-known Grad-CAM method \cite{selvaraju2017grad}. Grad-CAM uses gradients to compute the importance of spatial locations within convolutional layers. This allows us to visualize and interpret the differences in attention between models.

In this section, we present Grad-CAM maps under different conditions to better understand the effect of AttZoom:
\begin{itemize}
    \item We compare models with and without AttZoom layers. To avoid overloading the paper with visualizations, we focus on two models that show a clear performance gain with AttZoom (e.g., ResNet50 and ResNet50 CBAM) and two others with more modest improvements (e.g., DenseNet121 and ResNeXt), across different datasets. 
    \item We include two types of scenario: images where both models correctly classify the input (Scenario 1) and images where both models fail (Scenario 2). This avoids cherry-picking favorable examples.
\end{itemize}

Figure \ref{GradCam} shows the results. Scenario 1, where both models are correct, is the most informative. The key observation is that models with AttZoom layers are more fine-grained and have a diverse collection of details. A clear case is observed with fish images in TinyImageNet: while models without AttZoom focus on a single feature or a broad region, models with AttZoom attend to multiple smaller regions, such as the dorsal and lateral fins and the mouth.

In Scenario 2, where both models fail, neither is able to focus on the relevant regions or on enough details to produce a correct prediction. However, we still observe that models with our proposed AttZoom tend to explore more distinct areas of the image. This broader and finer attention is what underlies the performance improvements reported in Tables \ref{tab:AZCIFAR} and \ref{tab:AZIMAGENET}.

\subsection{Visual Analysis via Attention-Guided Warping}

In this section, we qualitatively illustrate the effect of integrating the AttZoom layer by spatially warping the image based on its attention. Specifically, we identify the regions where the AttZoom Layer focuses its attention and then apply a warping function to the input grid that expands high-attention areas and contracts low-attention ones.

Figure~\ref{warped} shows some of these warped examples. Here, we select correctly classified images and present one example per each of the 8 models introduced in Sect.~\ref{dataandmodels}, using Tiny ImageNet due to its higher-resolution samples. In the resulting visualizations, regions that are spatially expanded correspond to areas with strong attention responses, indicating that the model relies on those fine-grained details to make its predictions.

%% file: sec/7_conclusions.tex
\section{Conclusions} \label{sec:conclusions}

In this work, we proposed Attention Zoom (AttZoom): a spatial visual attention mechanism that allows networks to focus on finer image details, thereby enhancing performance. Unlike traditional attention mechanisms that rely on complex architectural changes, our method is modular, architecture-agnostic, and easy to integrate. This simplicity enables seamless incorporation into a wide range of architectures, including those that already employ attention mechanisms.

We demonstrated its effectiveness in various CNN backbones and datasets, showing consistent improvements in classification accuracy. Through both quantitative and qualitative analyses (including Grad-CAM visualizations and attention-based warping), we showed that models equipped with AttZoom attend to more informative and fine-grained regions in the input.

Our results suggest that AttZoom can serve as a plug-and-play module to improve visual attention in existing architectures with minimal parameter overhead. 

Future work will explore the application of AttZoom to other image processing and recognition tasks such as object detection and segmentation. Extensions of the AttZoom idea as a wrapper of existing AI models for application to text and temporal functions are also in our plans.

%% file: sec/8_acks.tex
\section*{Acknowledgement}
\label{Acknowledgement}

This work has been supported by projects BBforTAI (PID2021-127641OB-I00 MICINN/FEDER), HumanCAIC (TED2021-131787B-I00 MICINN), M2RAI (PID2024-160053OB-I00 MICIU/FEDER) and Cátedra ENIA UAM-VERIDAS (NextGenerationEU PRTR TSI-100927-2023-2). Work conducted in the ELLIS Unit Madrid. D. DeAlcala is supported by a FPU Fellowship (FPU21/05785). G. Mancera is supported by a FPI Fellowship (PRE2022-104499).

%% file: main.bbl
\begin{thebibliography}{59}
\providecommand{\natexlab}[1]{#1}
\providecommand{\url}[1]{\texttt{#1}}
\expandafter\ifx\csname urlstyle\endcsname\relax
  \providecommand{\doi}[1]{doi: #1}\else
  \providecommand{\doi}{doi: \begingroup \urlstyle{rm}\Url}\fi

\bibitem[Akiba et~al.(2019)Akiba, Sano, Yanase, Ohta, and Koyama]{optuna_2019}
Takuya Akiba, Shotaro Sano, Toshihiko Yanase, Takeru Ohta, and Masanori Koyama.
\newblock {Optuna: A Next-generation Hyperparameter Optimization Framework}.
\newblock In \emph{Proceedings of the 25th {ACM} {SIGKDD} International Conference on Knowledge Discovery and Data Mining}, 2019.

\bibitem[Anderson et~al.(2005)Anderson, Van~Essen, and Olshausen]{anderson2005directed}
Charles~H Anderson, David~C Van~Essen, and Bruno~A Olshausen.
\newblock {Directed Visual Attention and the Dynamic Control of Information Flow}.
\newblock \emph{Neurobiology of Attention}, pages 11--17, 2005.

\bibitem[Ba et~al.(2014)Ba, Mnih, and Kavukcuoglu]{ba2014multiple}
Jimmy Ba, Volodymyr Mnih, and Koray Kavukcuoglu.
\newblock {Multiple Object Recognition with Visual Attention}.
\newblock \emph{arXiv preprint arXiv:1412.7755}, 2014.

\bibitem[Brauwers and Frasincar(2023)]{survey3}
Gianni Brauwers and Flavius Frasincar.
\newblock {A General Survey on Attention Mechanisms in Deep Learning}.
\newblock \emph{IEEE Transactions on Knowledge and Data Engineering}, 35\penalty0 (4):\penalty0 3279--3298, 2023.

\bibitem[Carion et~al.(2020)Carion, Massa, Synnaeve, Usunier, Kirillov, and Zagoruyko]{carion2020end}
Nicolas Carion, Francisco Massa, Gabriel Synnaeve, Nicolas Usunier, Alexander Kirillov, and Sergey Zagoruyko.
\newblock {End-to-End Object Detection with Transformers}.
\newblock In \emph{Proceedings of the European conference on computer vision (ECCV)}, pages 213--229. Springer, 2020.

\bibitem[Chen et~al.(2020)Chen, Radford, Child, Wu, Jun, Luan, and Sutskever]{chen2020generative}
Mark Chen, Alec Radford, Rewon Child, Jeffrey Wu, Heewoo Jun, David Luan, and Ilya Sutskever.
\newblock {Generative Pretraining From Pixels}.
\newblock In \emph{Proceedings of the International Conference on Machine Learning (ICML)}, pages 1691--1703. PMLR, 2020.

\bibitem[Choromanski et~al.(2020)Choromanski, Likhosherstov, Dohan, Song, Gane, Sarlos, Hawkins, Davis, Mohiuddin, Kaiser, et~al.]{choromanski2020rethinking}
Krzysztof Choromanski, Valerii Likhosherstov, David Dohan, Xingyou Song, Andreea Gane, Tamas Sarlos, Peter Hawkins, Jared Davis, Afroz Mohiuddin, Lukasz Kaiser, et~al.
\newblock {Rethinking Attention with Performers}.
\newblock \emph{arXiv preprint arXiv:2009.14794}, 2020.

\bibitem[Chrabaszcz et~al.(2017)Chrabaszcz, Loshchilov, and Hutter]{chrabaszcz2017downsampled}
Patryk Chrabaszcz, Ilya Loshchilov, and Frank Hutter.
\newblock {A Downsampled Variant of ImageNet as an Alternative to the CIFAR datasets}.
\newblock \emph{arXiv preprint arXiv:1707.08819}, 2017.

\bibitem[Dai et~al.(2019{\natexlab{a}})Dai, Cai, Zhang, Xia, and Zhang]{dai2019second}
Tao Dai, Jianrui Cai, Yongbing Zhang, Shu-Tao Xia, and Lei Zhang.
\newblock {Second-Order Attention Network for Single Image Super-Resolution}.
\newblock In \emph{Proceedings of the IEEE/CVF Conference on Computer Vision and Pattern Recognition (CVPR)}, pages 11065--11074, 2019{\natexlab{a}}.

\bibitem[Dai et~al.(2019{\natexlab{b}})Dai, Yang, Yang, Carbonell, Le, and Salakhutdinov]{daietal2019transformer}
Zihang Dai, Zhilin Yang, Yiming Yang, Jaime Carbonell, Quoc Le, and Ruslan Salakhutdinov.
\newblock {Transformer-{XL}: Attentive Language Models beyond a Fixed-Length Context}.
\newblock In \emph{{Proceedings of the 57th Annual Meeting of the Association for Computational Linguistics}}, pages 2978--2988, Florence, Italy, 2019{\natexlab{b}}. Association for Computational Linguistics.

\bibitem[Daza et~al.(2024)Daza, Gomez, Fierrez, Morales, Tolosana, and Ortega-Garcia]{daza24att}
Roberto Daza, Luis~F. Gomez, Julian Fierrez, Aythami Morales, Ruben Tolosana, and Javier Ortega-Garcia.
\newblock {DeepFace-Attention}: Multimodal face biometrics for attention estimation with application to e-learning.
\newblock \emph{IEEE Access}, 12:\penalty0 111343--111359, 2024.

\bibitem[Delgado-Santos et~al.(2023)Delgado-Santos, Tolosana, Guest, Vera-Rodriguez, and Fierrez]{paula23gait}
Paula Delgado-Santos, Ruben Tolosana, Richard Guest, Ruben Vera-Rodriguez, and Julian Fierrez.
\newblock {M-GaitFormer}: Mobile biometric gait verification using transformers.
\newblock \emph{Engineering Applications of Artificial Intelligence}, 125:\penalty0 106682, 2023.

\bibitem[Delgado-Santos et~al.(2024)Delgado-Santos, Tolosana, Guest, Lamb, Khmelnitsky, Coughlan, and Fierrez]{paula24swipe}
Paula Delgado-Santos, Ruben Tolosana, Richard Guest, Parker Lamb, Andrei Khmelnitsky, Colm Coughlan, and Julian Fierrez.
\newblock {SwipeFormer}: Transformers for mobile touchscreen biometrics.
\newblock \emph{Expert Systems with Applications}, 237:\penalty0 121537, 2024.

\bibitem[Devlin et~al.(2019)Devlin, Chang, Lee, and Toutanova]{devlin2019bert}
Jacob Devlin, Ming-Wei Chang, Kenton Lee, and Kristina Toutanova.
\newblock {BERT: Pre-training of Deep Bidirectional Transformers for Language Understanding}.
\newblock In \emph{Proceedings of the Conference of the North American Chapter of the Association for Computational Linguistics: human language technologies, volume 1}, pages 4171--4186, 2019.

\bibitem[Fierrez(2006)]{fierrez06phd}
Julian Fierrez.
\newblock \emph{Adapted Fusion Schemes for Multimodal Biometric Authentication}.
\newblock PhD thesis, Univ. Politecnica de Madrid, 2006.

\bibitem[Fierrez et~al.(2018{\natexlab{a}})Fierrez, Morales, Vera-Rodriguez, and Camacho]{fierrez18fusion1}
Julian Fierrez, Aythami Morales, Ruben Vera-Rodriguez, and David Camacho.
\newblock Multiple classifiers in biometrics. {P}art 1: {F}undamentals and review.
\newblock \emph{Information Fusion}, 44:\penalty0 57--64, 2018{\natexlab{a}}.

\bibitem[Fierrez et~al.(2018{\natexlab{b}})Fierrez, Morales, Vera-Rodriguez, and Camacho]{fierrez18fusion2}
Julian Fierrez, Aythami Morales, Ruben Vera-Rodriguez, and David Camacho.
\newblock Multiple classifiers in biometrics. {P}art 2: {T}rends and challenges.
\newblock \emph{Information Fusion}, 44:\penalty0 103--112, 2018{\natexlab{b}}.

\bibitem[Fierrez-Aguilar et~al.(2005{\natexlab{a}})Fierrez-Aguilar, Garcia-Romero, Ortega-Garcia, and Gonzalez-Rodriguez]{fierrez05adapted}
J. Fierrez-Aguilar, D. Garcia-Romero, J. Ortega-Garcia, and J. Gonzalez-Rodriguez.
\newblock Adapted user-dependent multimodal biometric authentication exploiting general information.
\newblock \emph{Pattern Recognition Letters}, 26\penalty0 (16):\penalty0 2628--2639, 2005{\natexlab{a}}.

\bibitem[Fierrez-Aguilar et~al.(2005{\natexlab{b}})Fierrez-Aguilar, Ortega-Garcia, and Gonzalez-Rodriguez]{fierrez05score}
J. Fierrez-Aguilar, J. Ortega-Garcia, and J. Gonzalez-Rodriguez.
\newblock Target dependent score normalization techniques and their application to signature verification.
\newblock \emph{IEEE Trans. on Systems, Man \& Cybernetics}, 35\penalty0 (3):\penalty0 418--425, 2005{\natexlab{b}}.
\newblock Invited Paper.

\bibitem[Gonzalez-Sosa et~al.(2018)Gonzalez-Sosa, Vera-Rodriguez, Fierrez, and Ortega-Garcia]{ester18regions}
E. Gonzalez-Sosa, R. Vera-Rodriguez, J. Fierrez, and J. Ortega-Garcia.
\newblock Exploring facial regions in unconstrained scenarios: Experience on {ICB-RW}.
\newblock \emph{IEEE Intelligent Systems}, 33\penalty0 (3):\penalty0 60--63, 2018.

\bibitem[Gregor et~al.(2015)Gregor, Danihelka, Graves, Rezende, and Wierstra]{gregor2015draw}
Karol Gregor, Ivo Danihelka, Alex Graves, Danilo Rezende, and Daan Wierstra.
\newblock Draw: A recurrent neural network for image generation.
\newblock In \emph{{Proceedings of the International Conference on Machine Learning (ICML)}}, pages 1462--1471. PMLR, 2015.

\bibitem[Guo et~al.(2022)Guo, Xu, Liu, Liu, Jiang, Mu, Zhang, Martin, Cheng, and Hu]{guo2022attention}
Meng-Hao Guo, Tian-Xing Xu, Jiang-Jiang Liu, Zheng-Ning Liu, Peng-Tao Jiang, Tai-Jiang Mu, Song-Hai Zhang, Ralph~R Martin, Ming-Ming Cheng, and Shi-Min Hu.
\newblock {Attention mechanisms in computer vision: A survey}.
\newblock \emph{Computational Visual Media}, 8\penalty0 (3):\penalty0 331--368, 2022.

\bibitem[Hassanin et~al.(2024)Hassanin, Anwar, Radwan, Khan, and Mian]{HASSANIN2024102417}
Mohammed Hassanin, Saeed Anwar, Ibrahim Radwan, Fahad~Shahbaz Khan, and Ajmal Mian.
\newblock {Visual attention methods in deep learning: An in-depth survey}.
\newblock \emph{Information Fusion}, 108:\penalty0 102417, 2024.

\bibitem[He et~al.(2016)He, Zhang, Ren, and Sun]{he2016deep}
Kaiming He, Xiangyu Zhang, Shaoqing Ren, and Jian Sun.
\newblock {Deep Residual Learning for Image Recognition}.
\newblock In \emph{Proceedings of the IEEE/CVF Conference on Computer Vision and Pattern Recognition (CVPR)}, pages 770--778, 2016.

\bibitem[Hochreiter and Schmidhuber(1997)]{hochreiter1997long}
Sepp Hochreiter and J{\"u}rgen Schmidhuber.
\newblock Long short-term memory.
\newblock \emph{Neural Computation}, 9\penalty0 (8):\penalty0 1735--1780, 1997.

\bibitem[Howard(2017)]{howard2017mobilenets}
Andrew~G Howard.
\newblock {MobileNets: Efficient Convolutional Neural Networks for Mobile Vision Applications}.
\newblock In \emph{Proceedings of the IEEE/CVF Confernce on Computer Vision and Pattern Recognition (CVPR)}, pages 21--26, 2017.

\bibitem[Hsieh et~al.(2019)Hsieh, Lo, Chen, and Liu]{hsieh2019one}
Ting-I Hsieh, Yi-Chen Lo, Hwann-Tzong Chen, and Tyng-Luh Liu.
\newblock One-shot object detection with co-attention and co-excitation.
\newblock \emph{Advances in Neural Information Processing Systems}, 32, 2019.

\bibitem[Hu et~al.(2018)Hu, Shen, and Sun]{hu2018squeeze}
Jie Hu, Li Shen, and Gang Sun.
\newblock {Squeeze-and-Excitation Networks}.
\newblock In \emph{Proceedings of the IEEE/CVF Conference on Computer Vision and Pattern Recognition (CVPR)}, pages 7132--7141, 2018.

\bibitem[Hu et~al.(2020)Hu, Zhang, Jiang, Chaudhuri, Yang, and Nevatia]{hu2020span}
Xuefeng Hu, Zhihan Zhang, Zhenye Jiang, Syomantak Chaudhuri, Zhenheng Yang, and Ram Nevatia.
\newblock {SPAN: Spatial Pyramid Attention Network for Image Manipulation Localization}.
\newblock In \emph{Proceedings of the European conference on computer vision (ECCV)}, pages 312--328. Springer, 2020.

\bibitem[Huang et~al.(2017)Huang, Liu, Van Der~Maaten, and Weinberger]{huang2017densely}
Gao Huang, Zhuang Liu, Laurens Van Der~Maaten, and Kilian~Q Weinberger.
\newblock {Densely Connected Convolutional Networks}.
\newblock In \emph{Proceedings of the IEEE/CVF Conference on Computer Vision and Pattern Recognition (CVPR)}, pages 4700--4708, 2017.

\bibitem[Huertas-Tato et~al.(2022)Huertas-Tato, Martin, Fierrez, and Camacho]{camacho22cnns}
Javier Huertas-Tato, Alejandro Martin, Julian Fierrez, and David Camacho.
\newblock Fusing {CNNs} and statistical indicators to improve image classification.
\newblock \emph{Information Fusion}, 79:\penalty0 174--187, 2022.

\bibitem[Krizhevsky et~al.(2009)Krizhevsky, Hinton, et~al.]{krizhevsky2009learning}
Alex Krizhevsky, Geoffrey Hinton, et~al.
\newblock {Learning Multiple Layers of Features from Tiny Images}.
\newblock \emph{Toronto, ON, Canada}, 2009.

\bibitem[Lai et~al.(2020)Lai, Khan, Nie, Sun, Shen, and Shao]{lai2020understanding}
Qiuxia Lai, Salman Khan, Yongwei Nie, Hanqiu Sun, Jianbing Shen, and Ling Shao.
\newblock {Understanding More About Human and Machine Attention in Deep Neural Networks}.
\newblock \emph{IEEE Transactions on Multimedia}, 23:\penalty0 2086--2099, 2020.

\bibitem[Li et~al.(2020)Li, Du, Zhang, Wen, Luo, Wu, and Zhu]{li2020spatial}
Congcong Li, Dawei Du, Libo Zhang, Longyin Wen, Tiejian Luo, Yanjun Wu, and Pengfei Zhu.
\newblock {Spatial Attention Pyramid Network for Unsupervised Domain Adaptation}.
\newblock In \emph{Proceedings of the European Conference on Computer Vision (ECCV)}, pages 481--497. Springer, 2020.

\bibitem[Liu et~al.(2021)Liu, Lin, Cao, Hu, Wei, Zhang, Lin, and Guo]{liu2021swin}
Ze Liu, Yutong Lin, Yue Cao, Han Hu, Yixuan Wei, Zheng Zhang, Stephen Lin, and Baining Guo.
\newblock {Swin Transformer: Hierarchical Vision Transformer Using Shifted Windows}.
\newblock In \emph{Proceedings of the IEEE/CVF Conference on Computer Vision and Pattern Recognition (CVPR)}, pages 10012--10022, 2021.

\bibitem[Liu et~al.(2022)Liu, Mao, Wu, Feichtenhofer, Darrell, and Xie]{liu2022convnet}
Zhuang Liu, Hanzi Mao, Chao-Yuan Wu, Christoph Feichtenhofer, Trevor Darrell, and Saining Xie.
\newblock {A ConvNet for the 2020s}.
\newblock In \emph{Proceedings of the IEEE/CVF Conference on Computer Vision and Pattern Recognition (CVPR)}, pages 11976--11986, 2022.

\bibitem[Mancera et~al.(2025)Mancera, Morales, Fierrez, et~al.]{mancera2025pba}
Gonzalo Mancera, Aythami Morales, Julian Fierrez, et~al.
\newblock {PBa-LLM}: Privacy- and bias-aware {NLP} using named-entity recognition ({NER}).
\newblock In \emph{IAPR Intl. Conf. on Document Analysis and Recognition Workshops (ICDARw)}, 2025.

\bibitem[Mnih et~al.(2014)Mnih, Heess, Graves, and Kavukcuoglu]{mnih2014recurrent}
Volodymyr Mnih, Nicolas Heess, Alex Graves, and Koray Kavukcuoglu.
\newblock {Recurrent Models of Visual Attention}.
\newblock \emph{Advances in Neural Information Processing Systems}, 27, 2014.

\bibitem[Morales et~al.(2021)Morales, Fierrez, Vera-Rodriguez, and Tolosana]{morales21snets}
Aythami Morales, Julian Fierrez, Ruben Vera-Rodriguez, and Ruben Tolosana.
\newblock {SensitiveNets}: Learning agnostic representations with application to face recognition.
\newblock \emph{IEEE Trans. on Pattern Analysis and Machine Intelligence}, 43\penalty0 (6):\penalty0 2158--2164, 2021.

\bibitem[Navarro et~al.(2024)Navarro, Becerra, Daza, Cobos, Morales, and Fierrez]{becerra24eccvw}
Miriam Navarro, Álvaro Becerra, Roberto Daza, Ruth Cobos, Aythami Morales, and Julian Fierrez.
\newblock {VAAD}: Visual attention analysis dashboard applied to e-learning.
\newblock In \emph{European Conf. on Computer Vision Workshops (ECCVw)}, 2024.

\bibitem[Park et~al.(2018)Park, Woo, Lee, and Kweon]{park2018bam}
Jongchan Park, Sanghyun Woo, Joon-Young Lee, and In~So Kweon.
\newblock {Bam: Bottleneck Attention Module}.
\newblock \emph{arXiv preprint arXiv:1807.06514}, 2018.

\bibitem[Qin et~al.(2021)Qin, Zhang, Wu, and Li]{qin2021fcanet}
Zequn Qin, Pengyi Zhang, Fei Wu, and Xi Li.
\newblock {FcaNet: Frequency Channel Attention Networks}.
\newblock In \emph{Proceedings of the IEEE/CVF Conference on Computer Vision and Pattern Recognition (CVPR)}, pages 783--792, 2021.

\bibitem[Ren and Xia(2024)]{ren2024brain}
Jing Ren and Feng Xia.
\newblock {Brain-inspired Artificial Intelligence: A Comprehensive Review}.
\newblock \emph{arXiv preprint arXiv:2408.14811}, 2024.

\bibitem[Romero-Tapiador et~al.(2025)Romero-Tapiador, Tolosana, et~al.]{sergio25cvprw}
Sergio Romero-Tapiador, Ruben Tolosana, et~al.
\newblock Are vision-language models ready for dietary assessment? exploring the next frontier in {AI}-powered food image recognition.
\newblock In \emph{IEEE/CVF Conf. on Computer Vision and Pattern Recognition Workshops (CVPRw)}, 2025.

\bibitem[Selvaraju et~al.(2017)Selvaraju, Cogswell, Das, Vedantam, Parikh, and Batra]{selvaraju2017grad}
Ramprasaath~R Selvaraju, Michael Cogswell, Abhishek Das, Ramakrishna Vedantam, Devi Parikh, and Dhruv Batra.
\newblock {Grad-CAM: Visual Explanations From Deep Networks via Gradient-Based Localization}.
\newblock In \emph{Proceedings of the IEEE/CVF Confernce on Computer Vision and Pattern Recognition (CVPR)}, pages 618--626, 2017.

\bibitem[Tolosana et~al.(2021)Tolosana, Delgado-Santos, Perez-Uribe, Vera-Rodriguez, Fierrez, and Morales]{tolo21aaai}
Ruben Tolosana, Paula Delgado-Santos, Andres Perez-Uribe, Ruben Vera-Rodriguez, Julian Fierrez, and Aythami Morales.
\newblock {DeepWriteSYN}: On-line handwriting synthesis via deep short-term representations.
\newblock In \emph{AAAI Conf. on Artificial Intelligence (AAAI)}, pages 600--608, 2021.

\bibitem[Tolosana et~al.(2022)Tolosana, Romero-Tapiador, Vera-Rodriguez, Gonzalez-Sosa, and Fierrez]{tolo22fakes}
Ruben Tolosana, Sergio Romero-Tapiador, Ruben Vera-Rodriguez, Ester Gonzalez-Sosa, and Julian Fierrez.
\newblock Deepfakes detection across generations: Analysis of facial regions, fusion, and performance evaluation.
\newblock \emph{Engineering Applications of Artificial Intelligence}, 110:\penalty0 104673, 2022.

\bibitem[Tome et~al.(2013)Tome, Fierrez, Vera-Rodriguez, and Ramos]{tome13regions}
P. Tome, J. Fierrez, R. Vera-Rodriguez, and D. Ramos.
\newblock Identification using face regions: Application and assessment in forensic scenarios.
\newblock \emph{Forensic Science International}, \penalty0 (233):\penalty0 75--83, 2013.

\bibitem[Tome et~al.(2015)Tome, Fierrez, Vera-Rodriguez, and Ortega-Garcia]{tome15fusing}
Pedro Tome, Julian Fierrez, Ruben Vera-Rodriguez, and Javier Ortega-Garcia.
\newblock Combination of face regions in forensic scenarios.
\newblock \emph{Journal of Forensic Sciences}, 60\penalty0 (4):\penalty0 1046--1051, 2015.

\bibitem[Vaswani et~al.(2017)Vaswani, Shazeer, Parmar, Uszkoreit, Jones, Gomez, Kaiser, and Polosukhin]{vaswani2017attention}
Ashish Vaswani, Noam Shazeer, Niki Parmar, Jakob Uszkoreit, Llion Jones, Aidan~N Gomez, {\L}ukasz Kaiser, and Illia Polosukhin.
\newblock {Attention is All you Need}.
\newblock \emph{Advances in Neural Information Processing Systems}, 30, 2017.

\bibitem[Wang et~al.(2020)Wang, Wu, Zhu, Li, Zuo, and Hu]{Wang_2020_CVPR}
Qilong Wang, Banggu Wu, Pengfei Zhu, Peihua Li, Wangmeng Zuo, and Qinghua Hu.
\newblock {ECA-Net: Efficient Channel Attention for Deep Convolutional Neural Networks}.
\newblock In \emph{Proceedings of the IEEE/CVF Conference on Computer Vision and Pattern Recognition (CVPR)}, 2020.

\bibitem[Woo et~al.(2018)Woo, Park, Lee, and Kweon]{woo2018cbam}
Sanghyun Woo, Jongchan Park, Joon-Young Lee, and In~So Kweon.
\newblock {CBAM: Convolutional Block Attention Module}.
\newblock In \emph{Proceedings of the European Conference on Computer Vision (ECCV)}, pages 3--19, 2018.

\bibitem[Xie et~al.(2017)Xie, Girshick, Doll{\'a}r, Tu, and He]{xie2017aggregated}
Saining Xie, Ross Girshick, Piotr Doll{\'a}r, Zhuowen Tu, and Kaiming He.
\newblock {Aggregated Residual Transformations for Deep Neural Networks}.
\newblock In \emph{Proceedings of the IEEE/CVF Conference on Computer Vision and Pattern Recognition (CVPR)}, pages 1492--1500, 2017.

\bibitem[Xu et~al.(2015)Xu, Ba, Kiros, Cho, Courville, Salakhudinov, Zemel, and Bengio]{xu2015show}
Kelvin Xu, Jimmy Ba, Ryan Kiros, Kyunghyun Cho, Aaron Courville, Ruslan Salakhudinov, Rich Zemel, and Yoshua Bengio.
\newblock {Show, Attend and Tell: Neural Image Caption Generation with Visual Attention}.
\newblock In \emph{Proceedings of the International Conference on Machine Learning (ICML)}, pages 2048--2057. PMLR, 2015.

\bibitem[Yang et~al.(2018)Yang, Hao, Deng, Wei, Li, and Wang]{yang2018survey}
Shuangming Yang, Xinyu Hao, Bin Deng, Xile Wei, Huiyan Li, and Jiang Wang.
\newblock {A survey of brain-inspired artificial intelligence and its engineering}.
\newblock \emph{Life Research}, 1\penalty0 (1):\penalty0 23--29, 2018.

\bibitem[Yang et~al.(2020)Yang, Zhu, Wu, and Yang]{yang2020gated}
Zongxin Yang, Linchao Zhu, Yu Wu, and Yi Yang.
\newblock {Gated Channel Transformation for Visual Recognition}.
\newblock In \emph{Proceedings of the IEEE/CVF Conference on Computer Vision and Pattern Recognition (CVPR)}, pages 11794--11803, 2020.

\bibitem[Zhang et~al.(2022)Zhang, Wu, Zhang, Zhu, Lin, Zhang, Sun, He, Mueller, Manmatha, Li, and Smola]{Zhang_2022_CVPR}
Hang Zhang, Chongruo Wu, Zhongyue Zhang, Yi Zhu, Haibin Lin, Zhi Zhang, Yue Sun, Tong He, Jonas Mueller, R. Manmatha, Mu Li, and Alexander Smola.
\newblock {ResNeSt: Split-Attention Networks}.
\newblock In \emph{Proceedings of the IEEE/CVF Conference on Computer Vision and Pattern Recognition Workshops (CVPRw)}, pages 2736--2746, 2022.

\bibitem[Zhao and Wu(2019)]{zhao2019pyramid}
Ting Zhao and Xiangqian Wu.
\newblock {Pyramid Feature Attention Network for Saliency Detection}.
\newblock In \emph{Proceedings of the IEEE/CVF Conference on Computer Vision and Pattern Recognition (CVPR)}, pages 3085--3094, 2019.

\bibitem[Zhu et~al.(2020)Zhu, Su, Lu, Li, Wang, and Dai]{zhu2020deformable}
Xizhou Zhu, Weijie Su, Lewei Lu, Bin Li, Xiaogang Wang, and Jifeng Dai.
\newblock {Deformable DETR: Deformable Transformers for End-to-End Object Detection}.
\newblock In \emph{Proceedings of the International Conference on Learning Representations (ICLR)}, 2020.

\end{thebibliography}
